\documentclass[letterpaper, 10 pt, conference]{ieeeconf}

\IEEEoverridecommandlockouts
\usepackage[english]{babel}

\usepackage{subcaption}
\usepackage{tabularx}
\usepackage{amssymb}
\usepackage{amsmath}
\usepackage{stmaryrd}
\usepackage{bbold}
\usepackage{graphicx}
\usepackage{hyperref}
\hypersetup{colorlinks=true, allcolors=blue}
\usepackage{algorithm}
\usepackage{algpseudocode}
\usepackage{xcolor}
\usepackage{listings}

\usepackage{xparse}

\usepackage{cite}
\hypersetup{
  citecolor=black,
}
\newcommand{\kron}{\otimes}

\newcommand{\matr}[1]{\boldsymbol{#1}}

\newcommand{\zeros}[2]{
\ifthenelse{\equal{#2}{1}}{\vect{0}_{#1}}{\matr{\cancel{O}}_{#1 \times #2}}
}
\newcommand{\ones}[2]{
\ifthenelse{\equal{#2}{1}}{\vect{1}_{#1}}{\matr{1}_{#1 \kron #2}}
}
\newcounter{simulationcase}

\newcommand{\old}[1]{}  %comment not showed

\newcommand{\norm}[1]{\lVert#1\rVert}

\NewDocumentCommand{\codeword}{v}{%
\texttt{\textcolor{blue}{#1}}%
}

\title{\LARGE \bf
Control of Humanoid Robots with Parallel Mechanisms\\ using Differential Actuation Models
}

\author{Victor Lutz$^{1,*}$, Ludovic De Matteïs$^{1}$, Virgile Batto$^{1,2}$ Nicolas Mansard$^{1,3}$
\thanks{This work is suported by ROBOTEX 2.0 (ROBOTEX ANR-10-EQPX-0044 and TIRREX ANR-21-ESRE-0015), ANITI (ANR-19-P3IA-0004), by the French government (INEXACT ANR-22-CE33-0007 and "Investissements d'avenir" ANR-19-P3IA-0001) (PRAIRIE 3IA Institute), and by the Louis Vuitton ENS Chair on Artificial Intelligence}
\thanks{$^1$ Gepetto, LAAS-CNRS, Université de Toulouse, France}
\thanks{$^2$ Auctus, Inria, centre de l'université de Bordeaux, Talence, France}
\thanks{$^3$ Artificial and Natural Intelligence Toulouse Institute, France}
\thanks{$^*$ Corresponding author: \href{mailto:victor.lutz@laas.fr}{victor.lutz@laas.fr}}
}
\begin{document}
\maketitle
\thispagestyle{empty}
\pagestyle{empty}

\begin{abstract}
Several recently released humanoid robots, inspired by the mechanical design of Cassie, employ actuator configurations in which the motors are displaced from the joints to reduce leg inertia. 
While studies accounting for the full kinematic complexity have demonstrated the benefits of these designs, the associated loop-closure constraints greatly increase computational cost and limit their use in control and learning. 
As a result, the non-linear transmission is often approximated by a constant reduction ratio, preventing exploitation of the mechanism’s full capabilities.
This paper introduces a compact analytical formulation for the two standard knee and ankle mechanisms that captures the exact non-linear transmission while remaining computationally efficient. 
The model is fully differentiable up to second order with a minimal formulation, enabling low-cost evaluation of dynamic derivatives for trajectory optimization and of the apparent transmission impedance for reinforcement learning.
We integrate this formulation into trajectory optimization and locomotion policy learning, and compare it against simplified constant-ratio approaches. 
Hardware experiments demonstrate improved accuracy and robustness, showing that the proposed method provides a practical means to incorporate parallel actuation into modern control algorithms.
\end{abstract}
\section{Introduction}
\begin{figure}[!t]
\centering
\includegraphics[width=1\linewidth]{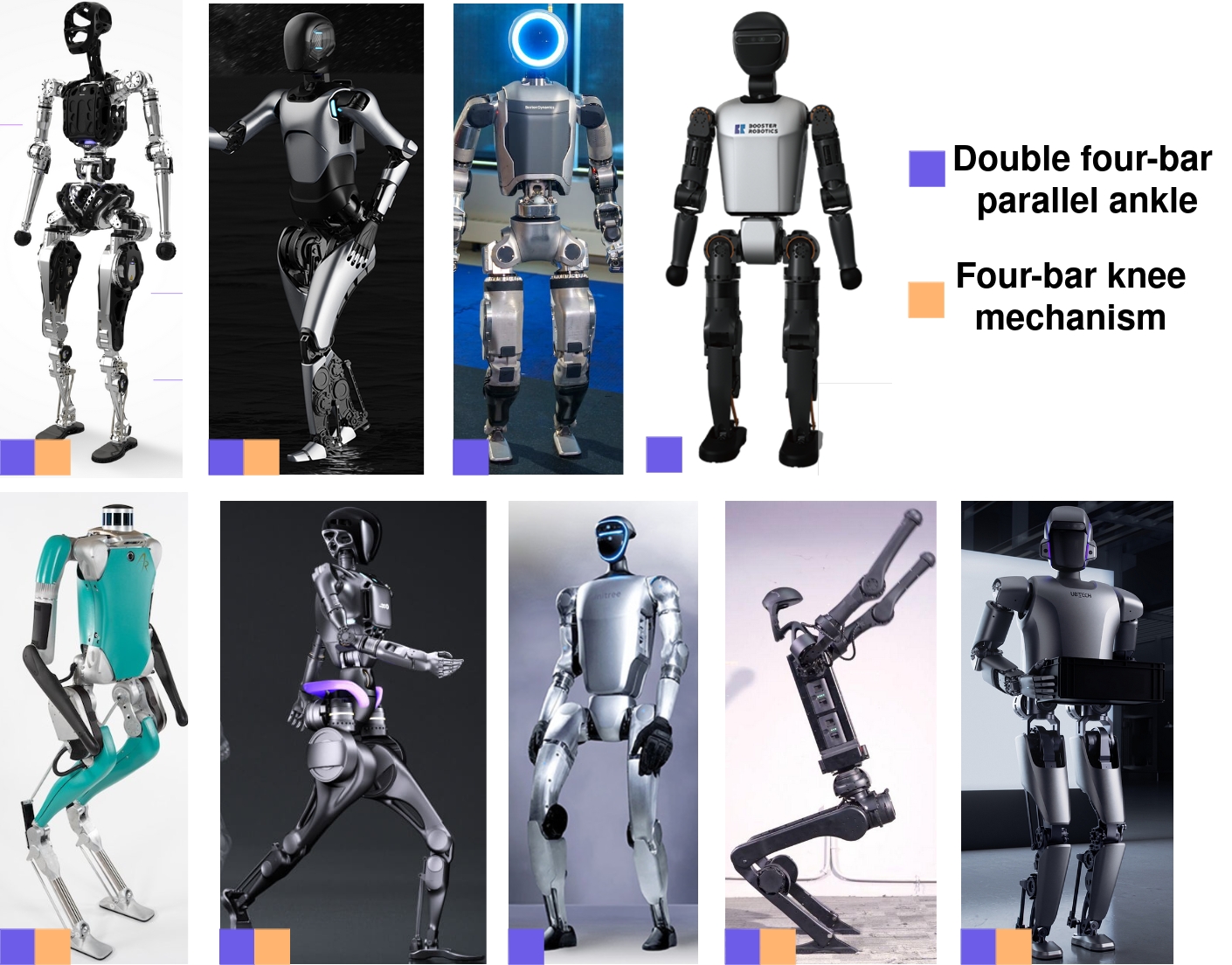}
\caption{Recent humanoids using parallel architectures. From top-left to bottom-right: Adam, A2, Atlas, T1, Digit V2, GR1, G1, H1, Walker S1 (Non exhaustive list)}
\label{Recent humanoids using parallel architectures}
\end{figure}

Recent advances in biped locomotion have largely been driven by new hardware designs.
Many companies and research groups, designing the most recent biped robots, shifted toward serial-parallel architecture (see Fig.~\ref{Recent humanoids using parallel architectures}), especially in the leg design.
These architectures reduce limb reflected inertia and improve impact absorption \cite{batto_comparative_2023, mronga_whole-body_2022}, thereby enabling more dynamic movements, albeit at the cost of more complex modeling and control. 

In most control frameworks - whether based on Whole-Body Model Predictive Control (WB-MPC) or Reinforcement Learning (RL) - motions are usually computed in the joint space using a serial dynamical model. 
% Previous work \cite{kumar_model_2019} has shown that computation of the dynamics with a serial model of this type could decrease the computation time without leading to significant errors, especially when side-chain inertias are kept fixed in the robot model . A low level layer, provided by the robot manufacturer, is then in charge of converting controls from the joint space to the actuator space at control time \cite{eser_design_2021}.  
% To mitigate the need for this conversion layer, some robot design rely on linear transmissions \cite{maslennikov_robust_2025, liao_berkeley_2024}.
% Previous works focused on a tighter modeling of the parallel mechanism itself, through analytical approaches \cite{zhou_comprehensive_2018, kumar_analytical_2019, hoffman_modeling_2024}. Others addressed motion generation using either reduced-dimensional (but non-serial) models \cite{liang_reduced-dimensional_2025} or full-dimensional models \cite{matteis_optimal_2024, mronga_whole-body_2022, sovukluk_whole_2023, boukheddimi_investigations_2023}.
% However, these last methods suffer from higher dimensionality and higher amount of constraints, leading to increased computational burden, making it difficult to run on real hardware. 
Several approaches have been described in the literature. 
The most direct approach is to exactly model the parallel branching of the kinematics, accounting for the closed-loop constraint and the additional internal DoF \cite{boukheddimi_investigations_2023}.
Although described early in the literature \cite{featherstone_rigid_2008}, the additional algorithmic cost has long been a bottleneck to their deployment until recent advances \cite{carpentier_proximal_2021, kumar_model_2019, matteis_optimal_2024, 10.1109/TRO.2024.3502515}.
Yet, the cost of solving the resulting motion problem remains.

On the other hand, dedicated analytical models have been formulated with reduced cost.
A systematic analytical model of the transmission is derived in \cite{kumar_analytical_2019}, from which we take inspiration in this work.
Dedicated studies have been proposed in the literature for some designs \cite{zhou_comprehensive_2018, kumar_analytical_2019, hoffman_modeling_2024}.
It has also been proposed to rely on heuristic models with equivalent dynamic property but reduced computational cost \cite{liang_reduced-dimensional_2025}.
On commercial robots, the handling of the parallel mechanism is often done in a closed-source algorithmic layer hidden to the user and sparsely documented, hence it is difficult to know what is implemented. To mitigate the need for this conversion layer, some robot design rely on unitary transmissions \cite{maslennikov_robust_2025, 11127524}.
Yet, we notice that most robots rely on kinematic mechanisms displaying a near-constant transmission ratio, which possibly allows to ignore the transmission non-linearity.
We will also show in the result section that such a hypothesis is quite limiting.

In RL, accounting for the full robot dynamics requires a simulator that supports closed-loop mechanisms. 
Until recently, such simulators were limited to CPU, making it challenging for locomotion policy learning \cite{li_reinforcement_2024}. 
Recent works used new generation GPU simulators such as Isaac Lab and Mujoco MJX to train locomotion policies with the full robot dynamics \cite{maslennikov_robust_2025, tanaka_mechanical_2025, amadio_learning_2025}. 
Despite some significant overhead, the computation time is only increased offline and does not impact policy inference.
However, these simulators yet rely on soft or approximate contacts models \cite{lidec_contact_2024} for constraining the closed chain, eventually increasing the sim-to-real gap and limiting deployments on real hardware.
This requires using specific techniques such as adversarial training or contact solver tuning to mitigate the drawbacks of the simulator \cite{maslennikov_robust_2025, tanaka_mechanical_2025, amadio_learning_2025}. A recent study has shown that policy training with a serial model and actuator-space output is possible \cite{zhang_lips_2025}. Yet, these methods come for now at extra costs and burden, are clearly not mature and fail to impact recent developments, which all rely on a limiting serial model. \\
In this paper, we propose a complete solution to accurately account for serial-parallel designs, both in WB-MPC and in RL, with a minimal additional implementation complexity and algorithmic cost.
Building on existing analytical models of the transmission \cite{kumar_analytical_2019}, we propose a dedicated formulation of the two most classical transmission (see Fig.~\ref{Recent humanoids using parallel architectures}) which leads to efficient derivatives.
These derivatives are key to the efficiency of WB-MPC solvers. 
Extending some early work \cite{8594112}, we show that these models modify the impedance of the actuator and derive an analytical model of it, also relying on the model derivatives.
Finally, we empirically evaluate the contributions and show the added value in the control compared to simplified serial model.

% To the best of our knowledge, the way the serial controls are transformed to actuator controls for RL policies is not documented in the literature, and is kept closed source by robot manufacturers. 
% Addressing this issue is one of the goals of this paper.

In the following, we first derive a \textbf{differential analytical model} of the geometric and kinematic mappings of the closed-loop transmissions composing modern robot architectures (Sec.~\ref{sec:actuation_model}) , obtaining cheap derivatives of the dynamics and the transmission reflected impedance.
We then propose two complete formulations to generate movements with this model, by trajectory optimization using the model derivatives and reinforcement learning using the derived transmission impedance (Sec. \ref{sec:implementation}).
The implementation of our contributions are empirically compared (Sec. \ref{sec:results}), demonstrating the benefits of such approach in TO and its capabilities through an RL framework with impedance transfer. We finally validate the method on a real robot with parallel actuation on both the knees and the ankles.
% We then propose a \textbf{Trajectory Optimization (TO)} formulation that leverages this differential actuation model to generate motions with serial dynamics that would otherwise be infeasible, especially under actuator constraints, providing all the necessary components to extend this framework to WB-MPC (Sec. \ref{subsec:OCP_description}).  
% We also implement a RL locomotion policy trained in serial space to a robot with non-linear actuation. (Sec. \ref{rldeploy}).  
% We first validate the effectiveness of our models in TO and then propose a RL framework along with a validation on a real biped robot.
%
% This paper will be structured as follows:
% First, in Sec.~\ref{sec:actuation_model}, we present a unified derivation of the geometric and kinematic mapping (so called Actuation models) for the knee and ankle mechanism and we compute their derivatives for their efficient use in both TO and RL, through impedance transfer.
% In Sec.~\ref{sec:implementation}, we then explain how we applied this method for TO (and how it can be extended to MPC) and for RL.
% Finally, in Sec.~\ref{sec:results}, we analyze the results of our implementations, demonstrating the benefits of such approach in TO and its capabilities through an RL framework with impedance transfer. We also validate the method on a real robot with parallel actuation on both the knees and the ankles.

\section{Method}
\label{sec:actuation_model}
% \begin{figure}
%     \centering
%     \includegraphics[width=1\linewidth]{img/model_comparaison.png}
%     \caption{Variation of closed-loop models. a) The \textit{Closed-Kinematics} models the entire chain and controls the motors, b) The \textit{Minimal Serial} model relies on an underlying serial chain and controls the serial joints torques, c) The \textit{Actuated Serial} model (ours) uses the same serial chain but control the serial joints torques through a kinematic actuation model.}
%     \label{Leg Model simplfication}
% \end{figure}
% We consider a robot composed of a main serial chain (carrying most of the inertia) whose motors actuate the joints through parallel linkages (See Fig.~\ref{Walk with our serial-parallel robot architecture} for an example of such architecture, also similar to H1 \cite{noauthor_unitree_sdk2exampleh1_nodate}, Adam \cite{AdamPnd} or Atlas \cite{ElectricAtlas}).

\begin{figure}
    \centering
    \includegraphics[width=0.6\linewidth]{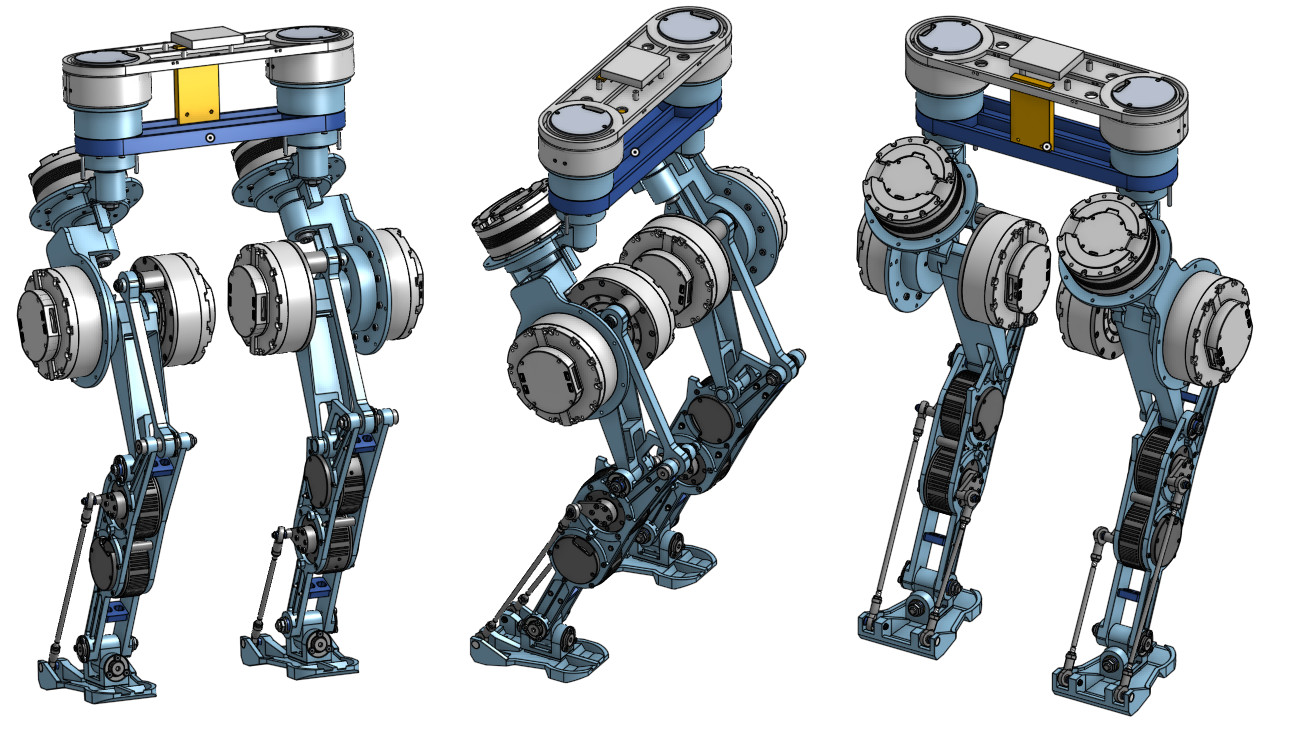}
    \caption{The Bipetto robot includes serial-parallel architecture, with parallel actuation for the knees and ankles.}
    \label{fig:bipetto}
\end{figure}

% While the most accurate model of a robot would consider the complete closed-loop kinematics, many of the results on recent robots have been obtained by neglecting the actuator mechanics, resulting in a simpler, although less accurate, model (in the latter we call it the \textit{Minimal Serial Model}).
% Rather than handling the complexity of the full model, we propose an intermediate solution, that we call \textit{Actuated Serial Model}, that accounts for the parallel linkage dynamics while being less expensive than a complete model.
In this section, we first provide the equations for a 1-DoF parallel transmission before extending it to a 2-DoF transmission.
These transmissions are typically used in the knee and ankle design of many robots. For clarity, we illustrate the designs with the robot Bipetto (Fig. \ref{fig:bipetto}), where the linkages clearly appear as it has no covers. However, all the developments directly apply on the other design shown in Fig \ref{Recent humanoids using parallel architectures}.
We then derive the expression to obtain the Actuation Jacobian (first order), relating the motor torque to the actuated serial joint torque, and the transmission dynamics (second order). 
Finally, we present a method to transfer impedance gains from the serial-space to the actuator-space, allowing impedance control and maintaining stabilization properties.

\subsection{Direct Geometry}
\subsubsection{Four-bar linkages} \label{subsec:fourbargeom}
\begin{figure}
\centering
    \begin{subfigure}{.5\linewidth}
      \centering
      \includegraphics[width=\linewidth, trim={0cm 0cm 0cm 9cm}, clip]{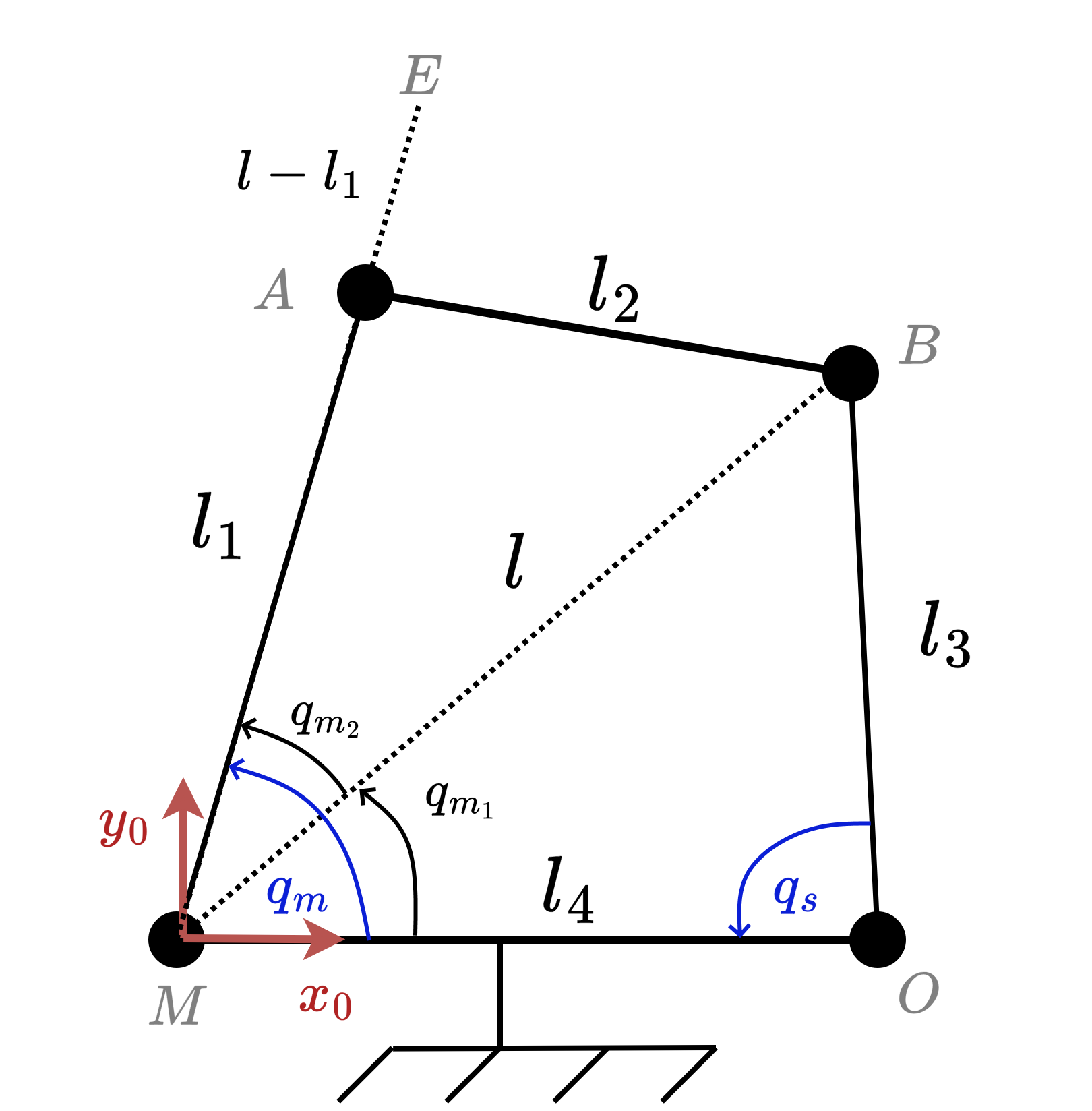}
    \end{subfigure}%
    \begin{subfigure}{.3\linewidth}
      \centering
      \includegraphics[width=\linewidth]{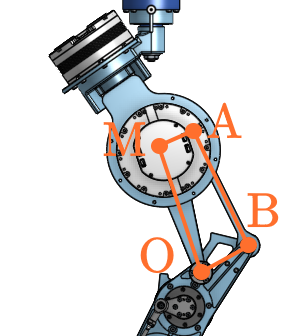}
    \end{subfigure}
    \caption{Planar 4-bar mechanism, with the serial link rotating around O, of angle $q_s$, motor rotating around M of angle $q_m$, B the attachment of the linkage on the lower limb and A the joint of the closed-loop linkage. A concrete example is given with the knee of the Bipetto robot.}
    \label{fig:planar-4-bar}
\end{figure}

We first consider a simple planar four-bars linkage transmission, present in many modern robots such as in the knee of H1 \cite{UnitreeH1} or in the ankle of Talos \cite{stasse2017talos}.
The knee actuation of our robot Bipetto is shown in Fig.~\ref{fig:bipetto} and Fig.~\ref{fig:planar-4-bar}. In this mechanism we denote $q_m$ the motor joint configuration and $q_s$ the configuration of the actuated serial joint.\\
The transmission is represented by the mapping $f$ for which we seek an efficient formulation.
\begin{equation}
    q_m \triangleq f(q_s)
\end{equation}
\noindent We denote by $b = (x_B, y_B)^T\in \mathbb{R}^2$ the coordinate vector of point B in the motor frame, $l(b) = \norm{b}$ its norm, $q_{m_1}=\widehat{OMB}$, $q_{m_2}=\widehat{BMA}$ and $r = cos(q_{m_2})$.
Applying Al-Kashi theorem (law of cosines) yields:
\begin{equation}
    r(b) = \frac{l(b)^2 + l_1^2  - l_2^2}{2l(b)l_1}
\end{equation}
The motor configuration is given by $q_m=q_{m_1} + q_{m_2}$, with:
\begin{equation}
\label{equ:qm1qm2expressions}
\left\{
\begin{aligned}
  &q_{m_1}(b) = atan2(y_B,x_B) \\
  &q_{m_2}(b) = acos(r(b))
\end{aligned}
\right.
\end{equation}
We restrict our derivations to $m_2 \in [0, \pi]$, giving a bijection between $r$ and $m_2$. We get that the motor configuration $q_m(b) = q_{m_1}(b) + q_{m_2}(b)$ is a function of the position of the point B.
Moreover, we can express the coordinate of the point B in the motor frame as a function of the serial joint configuration: $b(q_s) = [l_4  + l_3 \cos(q_s), l_3 \sin(q_s)]$, or directly express it as the forward kinematics of the serial chain denoted by $FK$:
\begin{equation}
    \label{equ:forward_kinematics_b}
    b(q_s) = FK(q_s)
\end{equation}
Combining \eqref{equ:forward_kinematics_b} with \eqref{equ:qm1qm2expressions} gives the complete expression of $q_m = f(q_s)$.

\subsubsection{Intricate four-bar linkages}
\begin{figure}
    \hspace{0.5cm}
    \begin{subfigure}{.55\linewidth}
        \centering
        \includegraphics[width=1.0\linewidth]{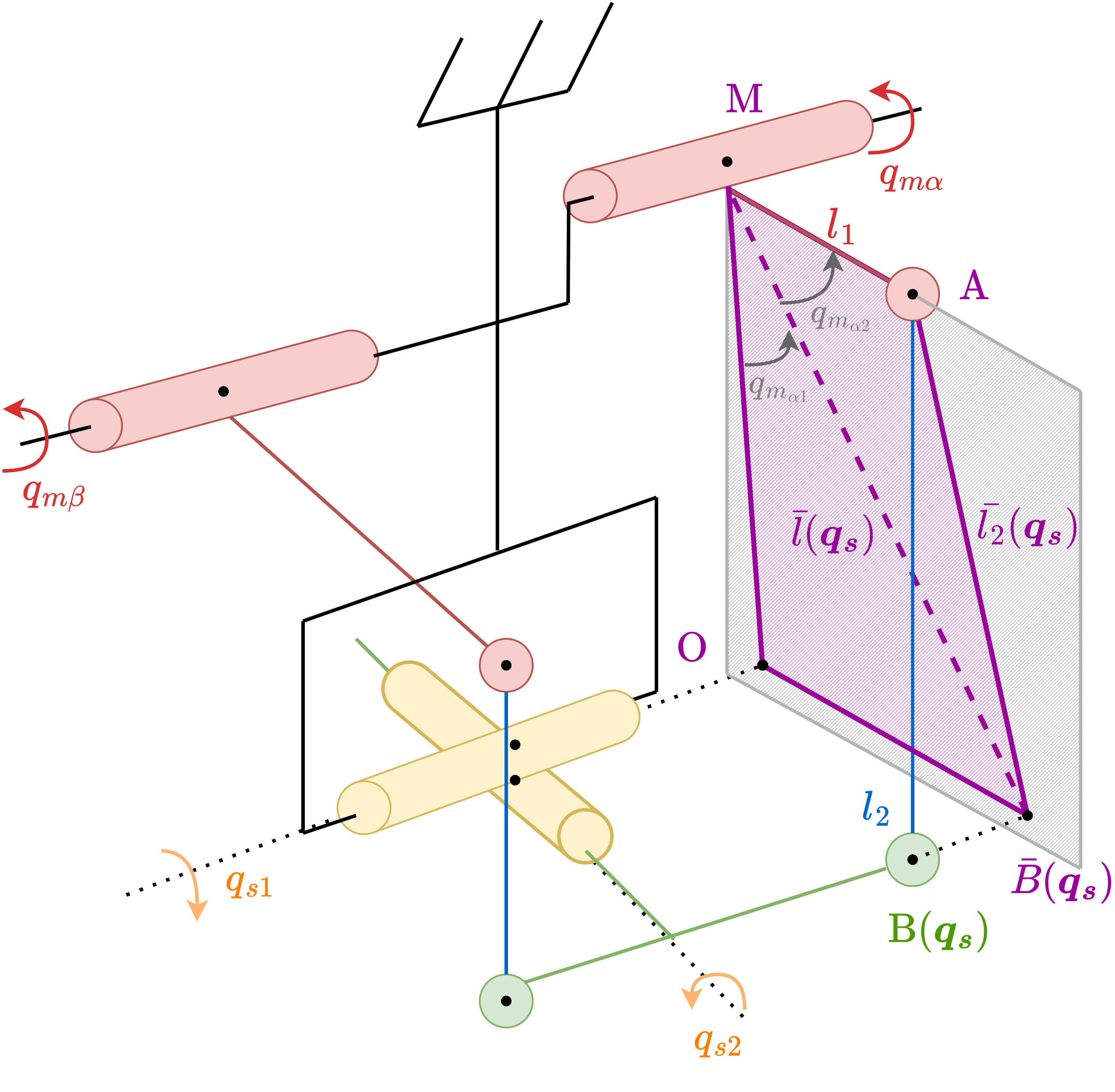}
    \end{subfigure}%
    \begin{subfigure}{.35\linewidth}
        \centering
        \includegraphics[width=1.0\linewidth,]{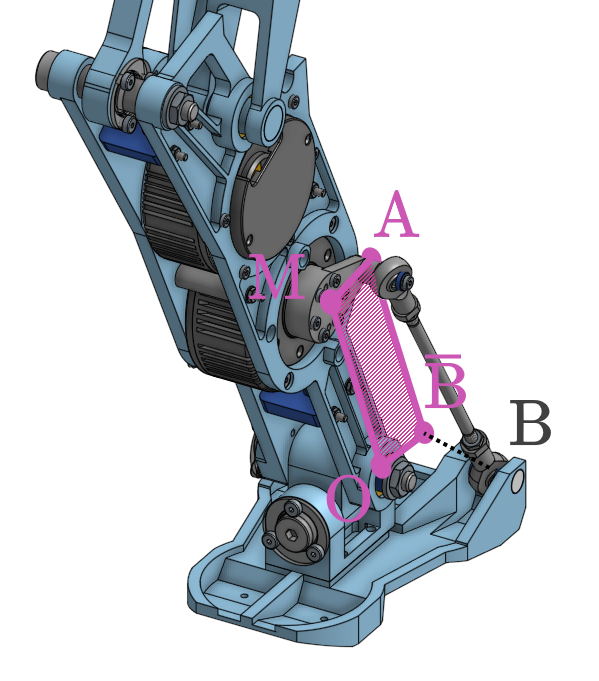}
    \end{subfigure}
    \caption{Sub projected planar four-bar from the ankle in purple. A concrete example is given on the right (ankle of the Bipetto robot)}
    \label{fig:ankle_mechanism}
\end{figure}

A more complex mechanism, consisting in two intricate four-bar linkages, yielding a coupled control of two serial DoF using two motors, can be observed in the ankles of H1, G1, GR1, Digit, T1.
A schematic representation of this mechanism is presented in Fig. \ref{fig:ankle_mechanism}.
While this mechanism is not planar, each side of it (denoted $\alpha$ and $\beta$) can be projected into a plane, orthogonal to the motor joint and going through the linkage point A, resulting in a virtual planar four-bars linkage.

Focusing on one side of the mechanism, we denote as $\bar B$ the projection of $B$ on the plane, so that $b \in \mathbb{R}^3$ and $\bar b = (x_B, y_B)^T \in \mathbb{R}^2$. 
Let us note, $l_1=\norm{\vec{AM}}$, $l_2=\norm{\vec{AB}}$, $\bar{l_2}^2(z_B) = l_2^2 - z_B^2=\norm{\vec{A\bar B}}$, and $\bar l(\bar b) = \norm{M \bar B}$. 
We can now proceed as before, with $r(b)$ that becomes:
\begin{equation}
    r(b) = \frac{\bar l(\bar b)^2 + l_1^2  - \bar{l_2}(z_B)^2}{2\bar l(\bar b) l_1}
\end{equation}

By doing so, we get a virtual planar four-bar mechanism that gives a relation between one motor configuration $q_{m_\alpha}$ and the ankle configuration $q_s = (q_{s_1} \quad q_{s_2})^T$.
Applying this method on both sides of the mechanism yields the relation:
\begin{equation}
    q_m = \begin{pmatrix}
        q_{m_\alpha} \\
        q_{m_\beta}
    \end{pmatrix} = \begin{pmatrix}
        f_{\alpha}(q_s) \\
        f_{\beta}(q_s)
    \end{pmatrix} = f(q_s)
\end{equation}
This relation generalizes the four-bar linkage (as it is its own projection) and we will use this notation for both mechanisms.

\subsection{Actuation Jacobian}
To transfer motor controls to their actions on the serial joints, we need a relation between the motor torques $\tau_m$ and the serial joints torques $\tau_s$.
To obtain such a relation, we compute the so called \textbf{Actuation Jacobian} $J_A$, function of $q_s$ that satisfies the relations:
\begin{equation}
\label{equ:japroperties}
\left\{
    \begin{aligned}
        \dot{q}_m &= J_A(q_s) \dot{q}_s \\
        \tau_s &= J_A(q_s)^T \tau_m
    \end{aligned}
\right.
\end{equation}

In this section, we will focus on the intricate four-bar mechanism. We first consider the effect of a single motor on the serial joints, before summing the contributions of the two motors (denoted $M_\alpha$ and $M_\beta$).
Without loss of generality, we derive the equations for $\tau_{m_\alpha}$ and denote $\tau_{s[\tau_{m_\alpha}]}$ the serial torques it induces and $J_{A, \alpha}$ the corresponding Actuation Jacobian.
The derivations for $\tau_{s[\tau_{m_\beta}]}$ do not add any additional complexity.
Computing the derivative of the mapping $f$ with respect to $q_s$ gives
\begin{equation}
     \frac{d q_{m_\alpha}}{d q_s} = \frac{d f_\alpha(q_s)}{d q_s} \triangleq J_{A,\alpha}(q_s)
\end{equation}
Since we focus on one side of the mechanism, we drop the subscript $\alpha$ from the four-bar notations, using for instance $q_{m_1}$ instead of $q_{m_{\alpha_1}}$.
To apply the chain rule, we write $q_{m_\alpha} = f_\alpha(q_s) = q_{m_1}(b) + q_{m_2}(r(\bar l(\bar b), \bar{l_2}(z_B)))$ and $b = \begin{pmatrix}
    \bar b \\ z_B
\end{pmatrix} = FK(q_s)$, yielding:
\begin{equation}
    \frac{dq_{m_\alpha}}{dq_s} = \left(\frac{dq_{m_1}}{db} + \frac{dq_{m_2}}{db}\right)\frac{d b}{d q_s} 
\end{equation}
% \begin{equation}
%     \frac{dq_{m_\alpha}}{dq_s} = \left(\frac{dq_{m_1}}{db} +\frac{dq_{m_2}}{dr}\left(\frac{dr}{dl}\frac{dl}{db} + \frac{dr}{d\bar{l_2}}\frac{d\bar{l_2}}{db}\right)\right)\frac{\partial b}{\partial q_s} 
% \end{equation}
We will note $\frac{db}{dq_s} = B_s$ and compute it using the analytical derivatives of the forward kinematics, implemented in Pinocchio \cite{carpentier_analytical_2018}.
The other terms are computed from Eq. \eqref{equ:qm1qm2expressions}:
\begin{equation}
\frac{dq_{m_1}}{db} = \frac{1}{\bar l^2}(-y_B\quad x_B\quad 0)  = b^T \begin{bmatrix} 0 & 1/\bar l^2 & 0 \\ -1/\bar l^2 & 0 & 0 \\ 0 & 0 &0\end{bmatrix}
\end{equation}
%Introducing the notation $r' = \frac{dr}{dl} = \frac{1}{l_1} - \frac{r}{l}$, we can directly compute the second term as
Introducing $\hat{r} = sin(q_{m_2})$, we also get:
\begin{equation}
\label{equ:dqm2_db}
\frac{dq_{m_2}}{db} = \left\{
\begin{aligned}
\frac{dq_{m_2}}{d\bar{b}} &= \frac{\partial q_{m_2}}{\partial r}\frac{\partial r}{\partial \bar l}\frac{d \bar l}{d\bar b} = \frac{l_1 r - \bar l}{\hat r \bar l^2 l_1} \bar{b}^T \\
\frac{dq_{m_2}}{dz_B} &= \frac{\partial q_{m_2}}{\partial r}\frac{\partial r}{\partial \bar{l_2}}\frac{d\bar{l_2}}{dz_B} = \frac{-z_B}{\hat r \bar l l_1}     
\end{aligned}
\right.
\end{equation}
This can be condensed as
\begin{equation}
\frac{dq_{m_\alpha}}{db} =\frac{dq_{m_1}}{db} + \frac{dq_{m_2}}{db} = b^T\underbrace{\begin{bmatrix} \mu &\nu &0 \\  -\nu & \mu & 0 \\ 0 & 0 & \xi \end{bmatrix}}_K
\end{equation}
where $\mu = \frac{r l_1 - \bar l}{\hat{r} \bar l^2 l_1}$, $\nu=\frac{1}{\bar l^2}$ and $\xi= \frac{-1}{\hat{r}\bar ll_1}$. 
It follows:
\begin{equation}
    \label{equ:expression_Ja}
    J_{A, \alpha}(q_s) \triangleq \frac{d q_{m_\alpha}}{d q_s} = b(q_s)^T K(r(q_s),l(q_s)) B_s(q_s)
\end{equation}
It is worth noting that for the intricate four-bar (i.e. the ankle transmission), $B_s \in \mathbb{R}^{3\times2}$, giving that $J_{A, \alpha}$ (same holds for $J_{A, \beta}$) is an element of $\mathbb{R}^{1\times2}$ while it is a scalar for the four-bar mechanism (and equals $J_A$). 
For the full ankle, we obtain 
\begin{equation}
    J_A = \begin{bmatrix}
        J_{A, \alpha} \\
        J_{A, \beta}
    \end{bmatrix} \in \mathbb{R}^{2\times2}
\end{equation}
% We then obtain the actuation model
% \begin{equation}
%     \label{equ:actuation_model}
%     \tau_s = J_A^T(q_s)\tau_m
% \end{equation}

This matrix can evidently be obtained from any other analytical model such as \cite{kumar_analytical_2019}. However, the resulting formulation is very compact, although trying to compare the algorithmic cost is likely useless. We need it for the next step, since any automatic differentiation we tried failed to produce a short and efficient formulation of the second-order terms.

\subsection{Actuation derivatives}
\label{subsec:actuation_derivatives}
% To use our \textit{Actuated Serial} model within a derivative-based motion generation algorithm (cf Sec. \ref{subsec:OCP_description}), we need the derivatives of the dynamics with respect to states and controls.
% While, the derivatives of the serial dynamics are well known \cite{shubhamsinghEfficientAnalyticalDerivatives2021}, those of the actuation model, with respect to the control $\frac{\partial \tau_s}{\partial u}$ and states $\frac{\partial \tau_s}{\partial x_s}$, are missing.
We now extend the derivatives to $\frac{\partial \tau_s}{\partial u}$ and $\frac{\partial \tau_s}{\partial x_s}$, which we need to compute the derivatives of the dynamic \cite{carpentier_analytical_2018}.
Our approach allows controlling directly the motor torques, giving $u = \tau_m$, through the relation \eqref{equ:japroperties}.
The derivative of the serial torques with respect to $u$ is therefore the Actuation Jacobian.\\
The state of the robot consists of the serial joints configuration and velocity $x_s = [q_s, \dot{q}_s]$.
The derivative of the serial torques with respect to the articular velocity directly yields:
\begin{equation}
    \frac{d \tau_{s[\tau_{m_\alpha}]}}{d \dot{q}_s} = 0
\end{equation}
Differentiating $\tau_s$ with respect to the configuration $q_s$ is more complex, since the product rule naturally involve tensorial elements.
We will split again the Actuation Jacobian into two parts $J_{A,\alpha}$ and $J_{A, \beta}$, and focus on the first term.
Let us differentiate this product by rewriting it differently:
\begin{equation}
\label{equ_taustauma}
  \tau_{s[\tau_{m_\alpha}]} = J_{A, \alpha}^T \tau_{m_\alpha} = B_s^T \lambda 
\end{equation}
where $\lambda = K^T b \tau_{m_\alpha}$ corresponds to the force applied by the motor $\alpha$ on the point B.
Deriving this relation gives
\begin{equation}
\label{equ_dtaustauma}
    \frac{d \tau_{s[\tau_{m_\alpha}]}}{d q_s} = \frac{d B_s^T}{d q_s}\lambda + B_s^T\frac{d \lambda}{d b}B_s
\end{equation}
In the first term, we denote $\frac{d B_s}{d q_s}$ as $B_{ss}$ and directly compute the term $B_{ss}^T \lambda$ using the spatial algebra implemented in Pinocchio \cite{carpentier_pinocchio_2019} to avoid manipulating tensors.\\
The second term is given by
\begin{equation}
\label{equ_dlambdadb}
    \begin{aligned}
        \frac{d \lambda}{d b} &= \frac{d K^T}{d b} b \tau_{m_\alpha} + K^T \tau_{m_\alpha} \\
    \end{aligned}
\end{equation}
where the derivative of $K\left(\bar l(b), r(q_{m_2}(b)), \hat r(q_{m_2}(b))\right)$ with respect to $b$, is given by:
\begin{equation}
    \frac{d K}{db} = \frac{\partial K}{\partial \bar l}\frac{d \bar l}{db} + \left(\frac{\partial K}{\partial r}\frac{dr}{d q_{m_2}} + \frac{\partial K}{\partial \hat{r}}\frac{d\hat{r}}{d q_{m_2}}\right) \frac{d q_{m_2}}{db}
\end{equation}
The matrix elements $\frac{\partial K}{\partial \bar l}$, $\frac{\partial K}{\partial r}$ and $\frac{\partial K}{\partial \hat{r}}$ are computed by differentiating $\mu(\bar l, r, \hat r)$, $\nu(\bar l)$ and $\xi(\bar l, \hat r)$ separately.
The resulting non-zero elements are:
\begin{equation}
    \begin{aligned}
        &\mu_{\bar l} = \frac{\partial \mu}{\partial \bar l} = \frac{\bar l-2rl_1}{\hat{r}\bar l^3l_1} \quad &\mu_{r} = \frac{\partial \mu}{\partial r} = \frac{1}{\hat{r}\bar l^2} \\
        &\mu_{\hat r} = \frac{\partial \mu}{\partial \hat r} = \frac{\bar l - rl_1}{\hat{r}^2\bar l^2 l_1} \qquad &\nu_{\bar l} =\frac{\partial \nu}{\partial \bar l} = \frac{-2}{\bar l^3}\\
        &\xi_{\bar l} = \frac{\partial \xi}{\partial \bar l} =\frac{1}{\hat r \bar l^2 l_1} \qquad &\xi_{\hat r} = \frac{\partial \xi}{\partial \hat r} = \frac{1}{\hat{r}^2 \bar l l_1}
    \end{aligned}
\end{equation}
Finally, the covector $\frac{dl}{db}$ and the scalars $\frac{dr}{d q_{m_2}}$ and $\frac{d \hat r}{d q_{m_2}}$ are computed to be:
\begin{equation}
    \begin{aligned}
        &\frac{d\bar l}{db} = \frac{\bar b}{\bar l} \qquad \frac{d r}{d q_{m_2}} = -\hat r \qquad \frac{d \hat r}{d q_{m_2}} = r
    \end{aligned}
\end{equation}
Using \eqref{equ:dqm2_db} for $\frac{d q_{m_2}}{db}$ concludes the derivation of $\frac{d \lambda}{db}$ and gives
\begin{equation}
    \frac{d \tau_{s[\tau_{m_\alpha}]}}{d q_s} = B_{ss}^T \lambda + B_s^T\left( \frac{dK}{db}^T b  + K^T \right) B_s \tau_{m_\alpha}
\end{equation}
% \begin{equation*}
% \begin{aligned}
%    K' &= 
%    \begin{bmatrix} 
%    \mu_l + \mu_{q_{m_2}} \frac{d q_{m_2}}{dl} & \nu_l & 0 \\ 
%    -\nu_l & \mu_l + \mu_{q_{m_2}} \frac{d q_{m_2}}{dl} & 0 \\ 
%    0 & 0 & \xi_{l} + \xi_{q_{m_{2}}} \frac{d q_{m_2}}{dl}
%    \end{bmatrix}  
% \end{aligned}
% \end{equation*}
% where
% \begin{equation}
% \begin{aligned}
%     &\mu_l = \frac{\partial \mu}{\partial l} = \frac{l-2rl_1}{\hat{r}l^3l_1} \qquad &\mu_{q_{m_2}} = \frac{\partial \mu}{\partial q_{m_2}} = \frac{rl-l_1}{\hat{r}^2l^2l_1} \\
%     &\nu_l =\frac{\partial \nu}{\partial l} = \frac{-2}{l^3} \qquad &\frac{d q_{m_2}}{dl} = \frac{rl_1-l}{\hat{r}ll_1} \\
%     &\xi_l = \frac{\partial \xi}{\partial l} =\frac{1}{\hat r l^2 l_1} \qquad &\xi_{q_{m_2}} = \frac{\partial \xi}{\partial q_{m_2}} = \frac{r}{\hat{r}^2 l l_1}
% \end{aligned}
% \end{equation}
% Substituting in \eqref{equ_dtaustauma} and \eqref{equ_dlambdadb}, we get:
% \begin{equation}
%      {\frac{d \tau_{s[\tau_{m_\alpha}]}}{dq_s}} = B_{ss}^T\lambda + B_{s}^T(\frac{1}{l}  K'^T bb^T + K^T)B_s \tau_{m_\alpha}
% \end{equation}

In the end, we can recover the dependency from each motor by summing their influence to the final derivative:
\begin{equation}
    {\frac{d \tau_s}{dq_s}} = \sum_{i} {\frac{d \tau_{s[\tau_{m_i}]}}{dq_s}}
\end{equation}
with $i\in \{\alpha\}$ for the four-bar and $i\in \{\alpha, \beta\}$ for the ankle transmission.
\subsection{Serial impedance transfer}\label{subsec:gains}
The classical approach for controlling robots (in particular nearly systematic in RL or for sample based MPC \cite{xue2024fullordersamplingbasedmpctorquelevel}) is to impose an active impedance to the serial joint through a PD controller whose reference $q^*$ is the output of the high level policy. 
This approach provides a compliant control law implemented at high frequency on the robot motors that stabilizes the system between two control steps $t_k$ and $t_{k+1}$:
\begin{equation}
    \label{equ_pdcontroller}
    \tau(t) = K_P(q^*(t_k)-q(t)) - K_D\ \dot{q}(t) \qquad t \in [t_k, t_{k+1}] 
\end{equation}
While having constant stiffness and damping at joint level is desirable for policy training, transferring this impedance control to the actuators of a robot with parallel mechanism is not straightforward and requires a transfer of the gains from the joint-space into the actuator-space. 
At time $t_k$, a policy trained in serial space gives a reference $q_{s}^*(t_k)$ that generates a torque $\tau_{s}^{*}(t_k)$ using \eqref{equ_pdcontroller}. In the case of a PD in serial joint space, we denote $K_{Ps}$ and $K_{Ds}$ the impedance gains of the serial joint.
Using the Actuation Jacobian, we get the corresponding desired motor torque $\tau_{m}^{*}(t_k) = J_A(q_s(t_k))^{-T} \tau_{s}^{*}(t_k)$. 

\begin{figure}
    \centering
    \includegraphics[width=\linewidth]{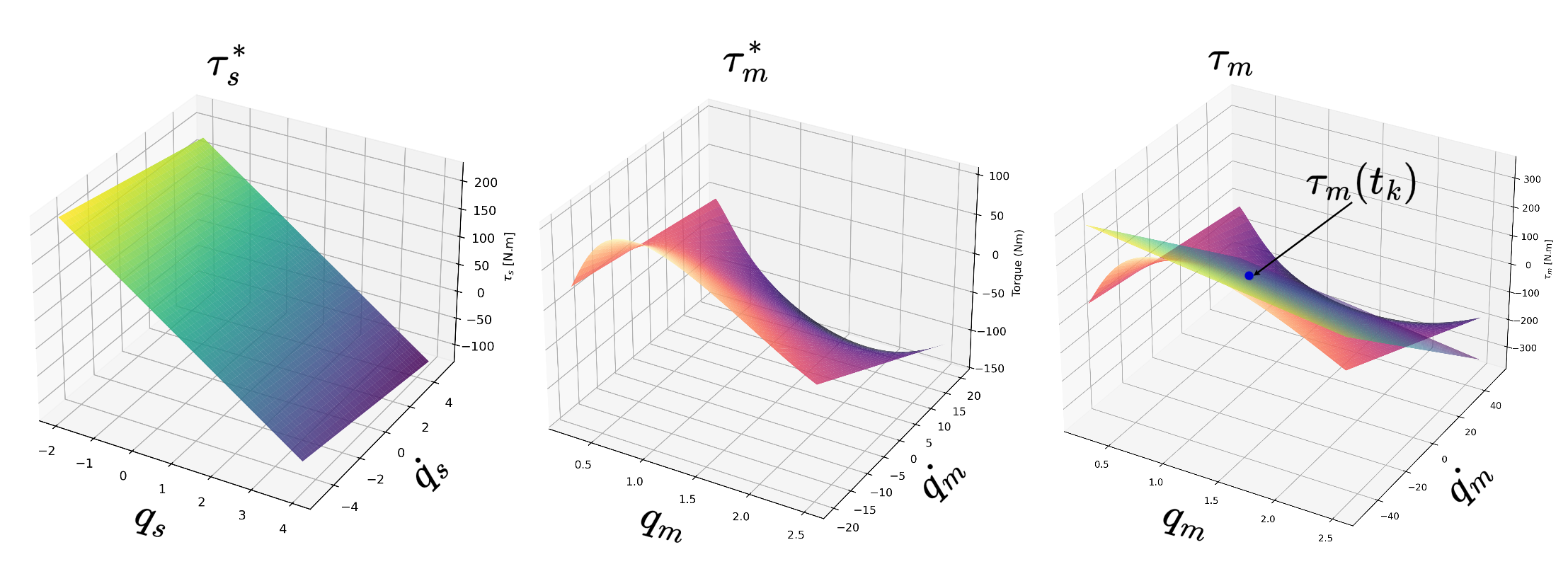}
    \caption{The constant serial gains generate an affine (with slopes $K_{Ps}$ and $K_{Ds}$) torque control in the joint-space (left). Computing the corresponding motor torques using the Actuation Model gives a non-linear control law (middle) that is approximated with a tangent plane (i.e. with correct $K_{Pm}$ and $K_{Dm}$) at the desired point $\tau_m^* = J_A^{-1}\tau_s^*$. Using the reference torque $\tau_m^*$ without feedback gains would lead to a horizontal plane which is a very wrong approximation of the curved manifold (in the middle). }
    \label{fig:variable_gains}
\end{figure}

To ensure the low-level controller acts in a stabilizing way, the derivatives of the applied torque with respect to the robot state must correspond to those of the desired torque (see Fig.~\ref{fig:variable_gains}). Using \eqref{equ_pdcontroller}, this condition at control time $t_k$ becomes \cite{8594112}:
\begin{equation}    
    \begin{aligned}
    \label{equ:KPmKdm_definition}
        \frac{d \tau_m}{dq_m}(t_k) = -K_{Pm}(t_k) = \frac{d \tau_{m}^{*}}{dq_m}(t_k) \\ 
        \frac{d \tau_m}{d \dot{q}_m}(t_k) = -K_{Dm}(t_k) = \frac{d \tau_{m}^{*}}{d \dot{q_m}}(t_k)
    \end{aligned}
\end{equation}
where $K_{Pm}(t_k)$ and $K_{Dm}(t_k)$ denote the impedance gains in the motor-space for $t \in [t_k, t_{k+1}]$. 
Developing \eqref{equ:KPmKdm_definition} gives:
\begin{equation}
    \begin{aligned}
    K_{Pm}^{*}(t_k) = &\underbrace{-(\frac{d J_A(q_s)^{-T}}{dq_m})}_{A_{Pm}} \tau_s + \underbrace{J_A(q_s)^{-T}K_{Ps}\frac{d q_s}{dq_m}}_{B_{Pm}} \\ &+ \underbrace{J_A(q_s)^{-T}K_{Ds}\frac{d \dot{q_s}}{dq_m}}_{C_{Pm}}
    \end{aligned}
\end{equation}

Previous studies \cite{8594112} adopted a similar approach, while neglecting the $A_{pm}$ and $C_{pm}$ terms by assuming that  $(q_{s}^{*} - q_s)$ remains small. However, this assumption is not valid in the context of reinforcement learning.
The term $B_{Pm}$ can directly be written as:
\begin{equation}
  B_{Pm} = J_A(q_s)^{-T}K_{Ps}J_A(q_s)^{-1}  
\end{equation}
Let us derive the $A_{Pm}$ term:
\begin{equation}
A_{Pm} =  J_A^{-T}(\frac{d J_A(q_s) }{dq_m})^T J_A^{-T} \tau_s = J_A^{-T}(\frac{d J_A(q_s) }{dq_s} \frac{d q_s }{dq_m})^T\tau_m
\end{equation}
Which is computed in the same way as the derivative of \ref{equ_taustauma}:
\begin{equation}
    A_{Pm} = J_A^{-T} J_A^{-T}\frac{d J_A(q_s)^{T}\tau_m}{dq_s}|_{\tau_m=const.}
\end{equation}
Finally, we can derive the $C_{pm}$ term with similar operations:
\begin{equation}
    C_{Pm} = -J_A(q_s)^{-T}K_{ds}J_A(q_s)^{-1} \frac{d J_A(q_s)u}{dq_s}|_{u=J_A(q_s)^{-2}  \dot{q_m}}
\end{equation}

For the damping gain, we denote the analog terms in the derivative $A_{Dm}$, $B_{Dm}$ and $C_{Dm}$. Among those, only $C_{Dm}$ is non-zero:
\begin{equation}
    C_{dm} = J_A(q_s)^{-T}K_{Ds}J_A(q_s)^{-1}
\end{equation}

Finally, to make sure the control law is consistent, we enforce $\tau_m(t_k)=\tau_{m}^{*}(t_k)$ by finding $q_{m}^{*}$ with:
\begin{equation}
\begin{aligned}
        q_{m}^{*}(t_k) = (K_{Pm}^{*}(t_k))^{-1}[\tau_{m}^{*}(t_k)  + K_{Dm}^{*}(t_k)\dot{q_m}(t_k) \\ + K_{Pm}^{*}(t_k) q_{m}(t_k)]
\end{aligned}
\end{equation}

\subsection{State estimation}\label{stateestimation}
In a MPC or RL setting, the actual $q_s$ angles from the robot are needed.
However, on the real system, the motor encoders provide only $q_m$, requiring the inverse mapping $f^{-1}$ to recover $q_s$.
We can estimate $q_s$ via numerical optimization:
\begin{equation} 
q_s = f^{-1}(q_m) = \min_{\hat{q_s}} \mathcal{L}(\hat{q_s}, q_m) = \min_{\hat{q_s}} || f(\hat{q_s}) - q_m ||^2 
\end{equation}
Gradient of the state estimation cost function is given by 
\begin{equation}
    {\nabla \mathcal{L}}_{\hat{q_s}} = 2 J_A^T(\hat{q_s})( f(\hat{q_s}) - q_m) 
\end{equation}

State continuity allows using accurate warm starts, enabling rapid convergence of the minimization problem. Serial velocities $\dot{q_s}$ can be simply recovered using \ref{equ:japroperties}.

\section{Implementation}
\label{sec:implementation}
\subsection{Robot model}
\begin{figure}
\centering
    \begin{subfigure}{.33\linewidth}
       \centering
       \includegraphics[width=1.0\linewidth, trim={0cm 0.5cm 0cm 0cm}]{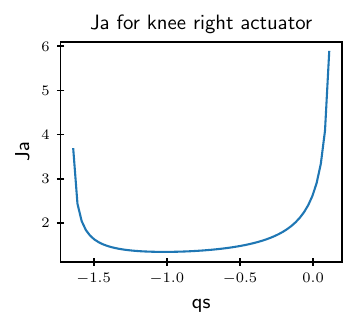}
       \caption{Knee transmission}
    \end{subfigure}%
    \begin{subfigure}{.55\linewidth}
      \centering
      \includegraphics[width=1.0\linewidth, trim={0cm -1.5cm 0cm 0cm}, clip]{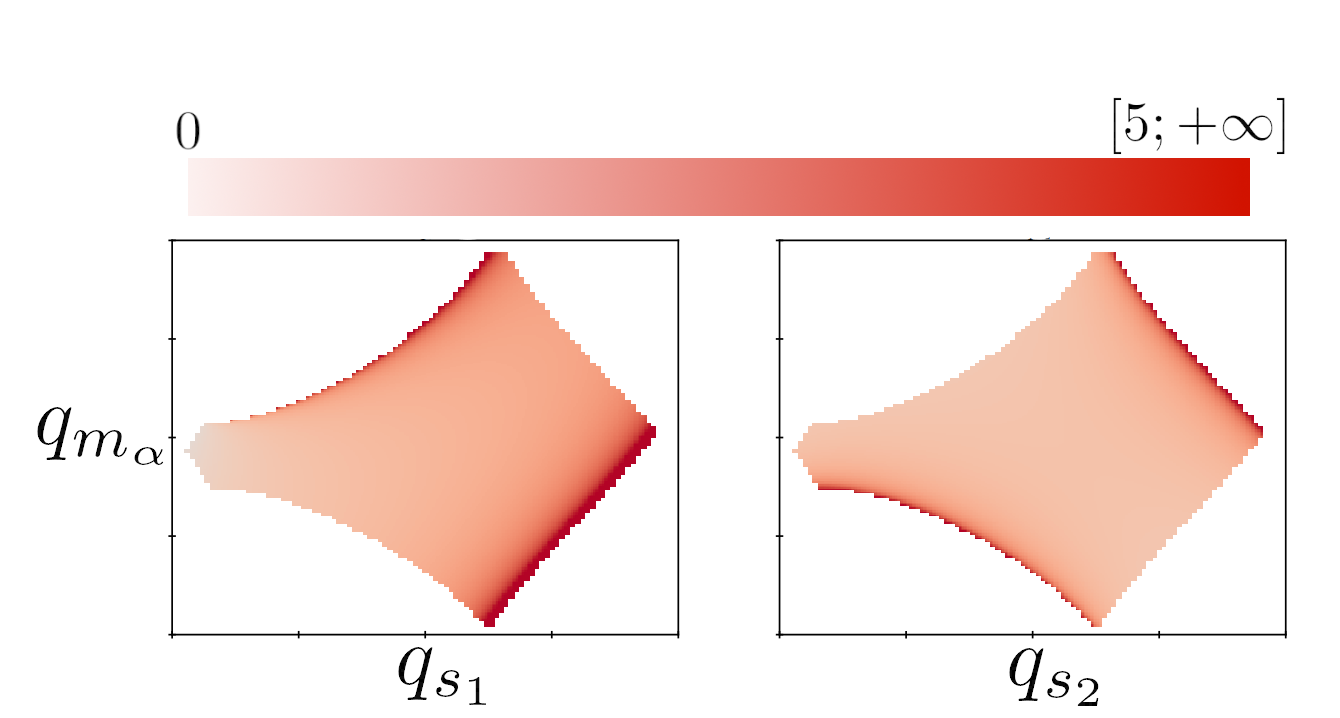}
      \caption{Ankle transmission}
    \end{subfigure}
    \caption{Variable transmission ratio $J_A(q_s)$ for parallel transmissions of our robot (four-bar and intricate-four bar). For the ankle, the contribution of motor $\alpha$ alone on the two DoF of the ankle - $q_{s_1}$ and $q_{s_2}$ - is shown ($\frac{d q_s}{d q_{m_\alpha}}$). The diamond shape on each plot correspond to the feasible space of the mechanism and the color represents the absolute value of the scalar reduction ratios $\frac{d q_{s_1}}{d q_{m_\alpha}}$ and $\frac{d q_{s_2}}{d q_{m_\alpha}}$.}
    \label{ja}
\end{figure}

We implemented our strategy on the Bipetto robot, showcased in Fig. \ref{fig:bipetto}.
% \footnote{Robot model at \textit{Link removed during the double-blind review}}
The robot is composed of a four-bar transmission for the knee joint and of an intricate four-bars transmission for the ankle, following the inspiration of Digit, H1 and others. 
We define the \textit{Minimal Serial} model of the robot by freezing the joints in the parallel transmission and adding fictive actuation in the serial joints, resulting in a model with 6 DoF per leg.\\
To account for the parallel actuation of the robot, represented in Fig.~\ref{fig:planar-4-bar} and Fig.~\ref{fig:ankle_mechanism}, we compute the corresponding actuation jacobians and their derivatives. 
This extends the \textit{Minimal Serial} model to create our \textit{Actuated Serial} model. 
A visual representation of the Actuation Jacobians for our robot architecture are shown in Fig.~\ref{ja}.

Note that the variable reduction ratio for each mechanism increases (in absolute value) when the parallel mechanism gets closer to singularities. For the knee actuation, this corresponds to fully bent and fully stretched legs.

\subsection{Trajectory Optimization}
\label{subsec:OCP_description}
\begin{figure}
    \centering
    \includegraphics[width=\linewidth]{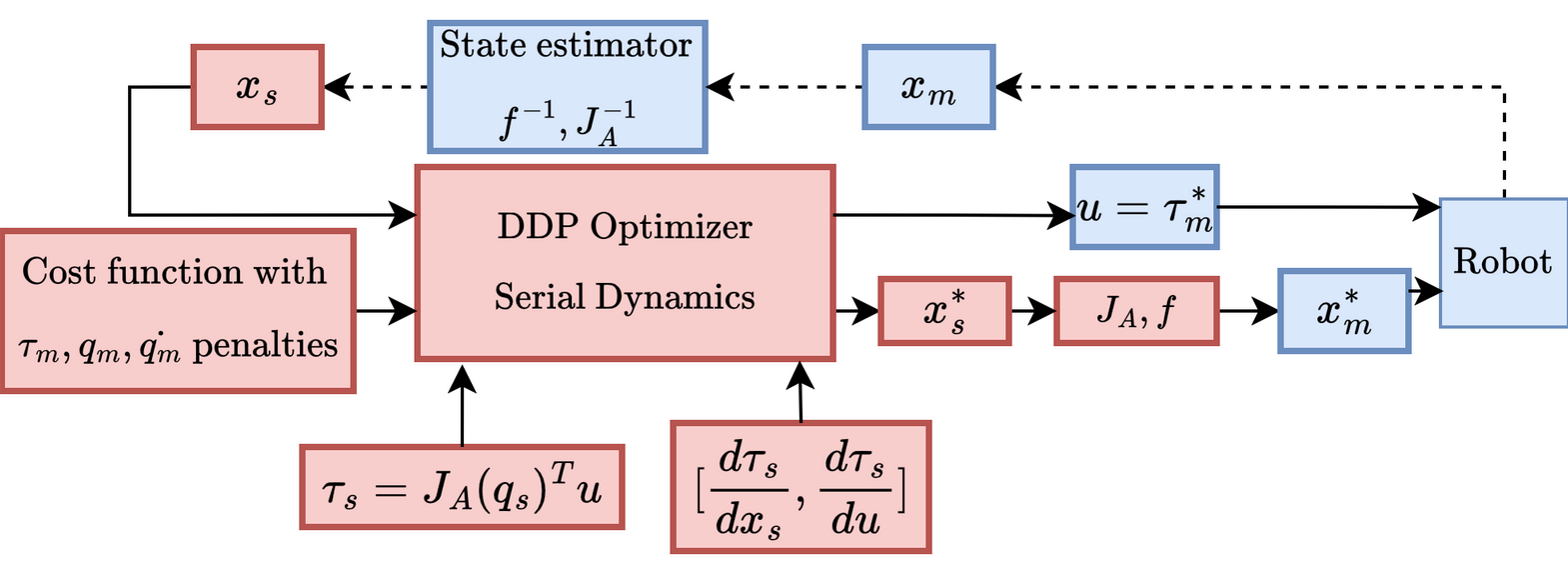}
    \caption{Whole Body Trajectory Optimization strategy. The OCP is solved with a serial dynamics controlled directly with motor torques, thanks to our derivatives of the Actuation Model. The problem can include cost and constraint on either serial states $x_s$ or motor states $x_m$. It outputs motor torques $\tau_m$ directly sent to the robot. The dotted line shows the extension needed to turn the TO into MPC.}
    \label{fig:TO_schematic}
\end{figure}

We formulate our control problem as a multiple shooting Optimal Control Problem (OCP), solving for controls $u[k]$ and states $x[k]$ at each time step $k$ \cite{diehl2006fast}. 
This classically transcribes as a Non-Linear Program (NLP)
\begin{equation} 
\label{equ:OCP_multiple_shooting_base}
    \begin{aligned}
        \min_{{U},\ {X}} & \sum_{k=0}^{N-1} l_k({x}[k], {u}[k]) + l_N({x}[N]) \\
        \text{s.t.} & \quad \forall k \in \llbracket 0, N-1\rrbracket \quad {x}{[k+1]} = f_k({x}[k], {u}[k]) \\
        % & \quad \forall k \in \llbracket 0, N-1\rrbracket \quad {c}_k({x}[k], {u}[k]) \leq {0} \\
        % & \quad {x}[0] = {x}_0
    \end{aligned}
\end{equation}
where $l_k$ defines the running costs, $l_N$ the terminal cost, and $f_k$ is the dynamics of the system. 
We opt for a simplified model of the robot dynamics by using a \textit{Minimal Serial} model with joint actuation defined as $\tau_s[k] = J_A^T u[k]$, so that the controls correspond to motor torques $u[k] = \tau_m[k]$ (see Fig.~\ref{fig:TO_schematic} for a global view of the TO strategy).
This formulation accounts for the parallel actuation nonlinearities through the Actuation Jacobian while keeping the model complexity as low as possible.\\
The problem \eqref{equ:OCP_multiple_shooting_base} is solved using the FDDP algorithm implemented within the Crocoddyl library \cite{mastalli_crocoddyl_2020}, and relies on Pinocchio \cite{carpentier_pinocchio_2019} for the serial dynamics computation.
% Thanks to our approach, the joint torques used in the dynamic computation now depend on the controls $u[k]$ through the actuation model described before, and yielding a variable reduction ratio.
Our formulation of the dynamics also allows setting constraints (such as limits) on motor torques directly, which would otherwise not be possible for the coupled motors of the ankle. \\
FDDP being a derivative based algorithm, it requires the derivatives of the dynamics with respect to the states and controls.
While the algorithms implemented in Pinocchio can be used to compute the derivatives of the serial dynamics $f_k(x[k], \tau_s[k])$ with respect to $x[k]$ and $\tau_s[k]$, it is not sufficient for our implementation since the dependencies $\tau_s(x[k], u[k])$ are not accounted for. Through our actuation model, we get $\tau_s[k] = \tau_s(x[k], u[k])$ and add the corresponding additional derivatives, that we derived in section \ref{subsec:actuation_derivatives}.\\
Using the actuation model also allows using costs and constraints directly in the motor-space even though the OCP uses serial states. For instance, we set in problem~\ref{equ:OCP_multiple_shooting_base} some lower and upper bounds ($\underline{q_m}$ and $\overline{q_m}$ respectively) on the ankle motor joints through the $f$ mapping:
\begin{equation}
    \underline{q_m} \leq f(q_s) \leq \overline{q_m}
\end{equation}
% Because of the parallel actuation, the feasible region in the configuration space of our system is not a box neither for the serial joints nor the motor joints. 
% However, as shown in Fig.\ref{q_s trajectories}, setting box limits on the ankle motor $q_m$ permits to reach a larger range of motion than applying box limits directly on the serial joints.
% Therefore, we set in \eqref{equ:OCP_multiple_shooting_base} some lower and upper bounds ($\underline{q_m}$ and $\overline{q_m}$ respectively) on the ankle motor joints through the $f$ mapping:
% \begin{equation}
    % \underline{q_m} \leq f(q_s) \leq \overline{q_m}
% \end{equation}

We implemented our approach for a walk on a flat and on a rough terrain, and for a stair climbing problem.
These tasks were chosen to push the robot close to his joint limits to demonstrate the impact of the non-linear transmission. 
The cost functions of \eqref{equ:OCP_multiple_shooting_base} are standard for this kind of problem and we refer the reader to \cite{dantec_whole-body_2022} for more details.

We compare the results with a \emph{Minimal Serial} model, where $u=\tau_s$ is turned a posteriori - when it is feasible - into motor torques using the relation $\tau_m = {J_A^{-T}} \tau_s$.

\subsection{Reinforcement Learning}\label{rldeploy}

\begin{figure}
    \centering
    \includegraphics[width=0.7\linewidth]{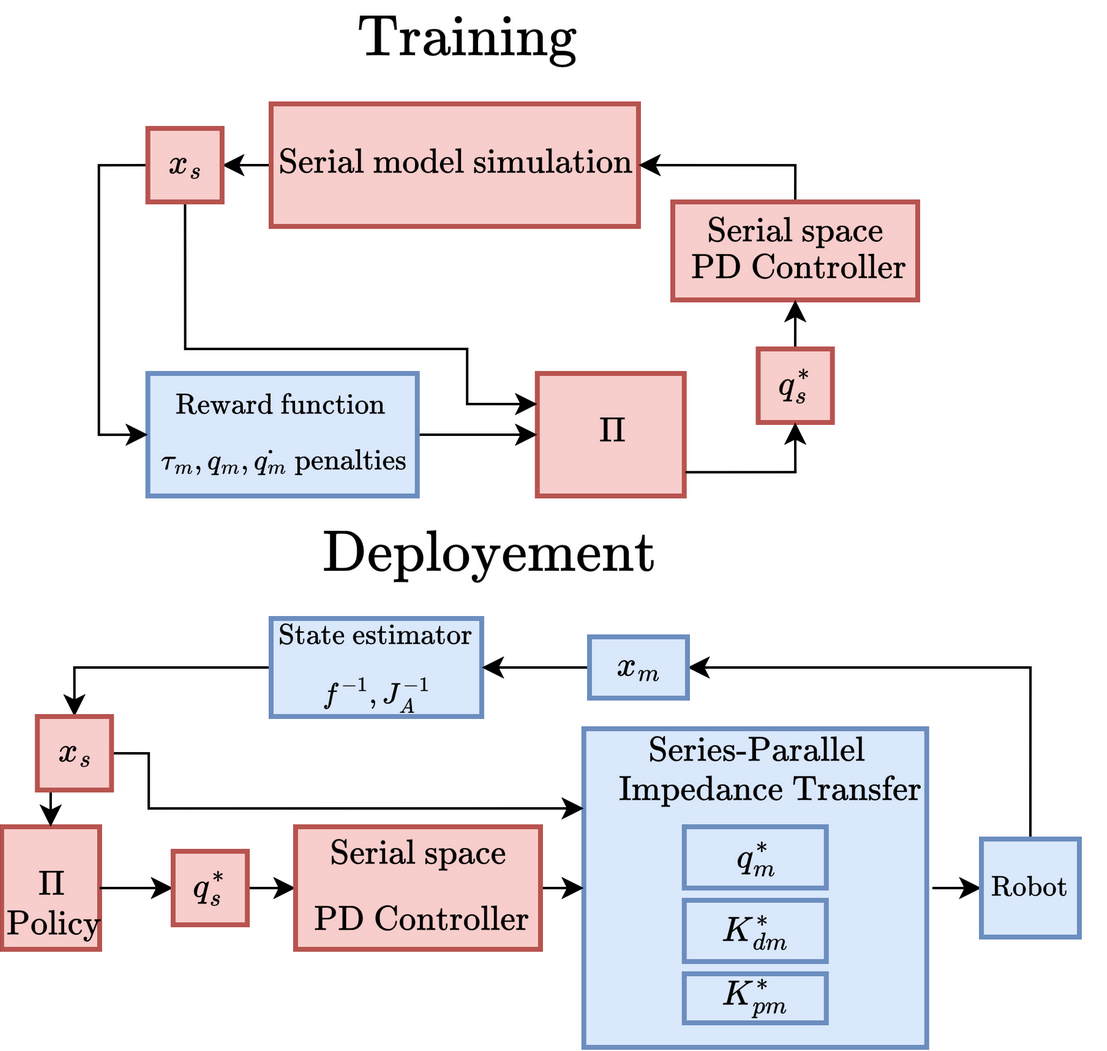}
    \caption{RL training and deployment. During deployment, serial references pass through a conversion step that outputs $q_{m}^{*}$, $K_{Pm}$ and $K_{Dm}$ for the actuator controller.}
    \label{rlmethod}
\end{figure}

We implemented bipedal locomotion in RL using the Isaac Lab framework. We used the constrained PPO formulation CaT \cite{chane-sane_cat_2024} to avoid extensive tuning. 
Reward design is outside the scope of this paper, however, for interested readers, our rewards and constraints formulation is very close to \cite{roux2025constrainedreinforcementlearningunstable}. 
Fig.~\ref{rlmethod} presents an overview of the training and deployment method, that we detail in the following sections.
Our policy was trained on a Nvidia RTX 4090 GPU with 4096 parallel environments.

\subsubsection{Observation space}
To use a serial model in the simplest way, our observations contains the serial joints positions $q_s$ and velocities $\dot{q_s}$, obtained through the state estimator described in \ref{stateestimation}. 
Our observation vector also contains the projected gravity vector, as well as the robot base angular velocity, and velocity tracking commands. All observations are stored during a history of 6 steps and then concatenated in the final observation vector. 

\subsubsection{Action space}
The policy is trained to output a serial position offset $\delta_{q_s}$ that is added to a reference standing position $q_{s0}$, to create the reference position $q_{s}^{*} = q_{s_0}+\delta_{q_s}$. 
This position offset then goes into a low-level simulated serial joint PD controller that generates torque $\tau_s$ for the joint at a higher rate: $\tau_s = K_{Ps} (q_{s}^{*} - q_s) - K_{Ds} \dot{q_s}$. 
The impedance gains $K_{Ps}$ and $K_{Ds}$ are chosen to be constant for simplicity.

\subsubsection{Series-parallel deployment}
Once trained, our policy is deployed on an embedded CPU and uses the impedance transfer described in section \ref{subsec:gains} to transfer the impedance gains from the serial-space to the actuator-space.
The policy inference is performed at $30$Hz, the reference $q_m^*$ and the gains $K_{Pm}$ and $K_{Dm}$ are updated at $100$Hz to better fit the curved manifold of Fig. \ref{fig:variable_gains} and the motors low-level impedance controller runs at about $1$kHz.

\section{Results}
\label{sec:results}
\subsection{Trajectory Optimization}

\begin{figure}
    \centering
    \includegraphics[width=0.5\linewidth]{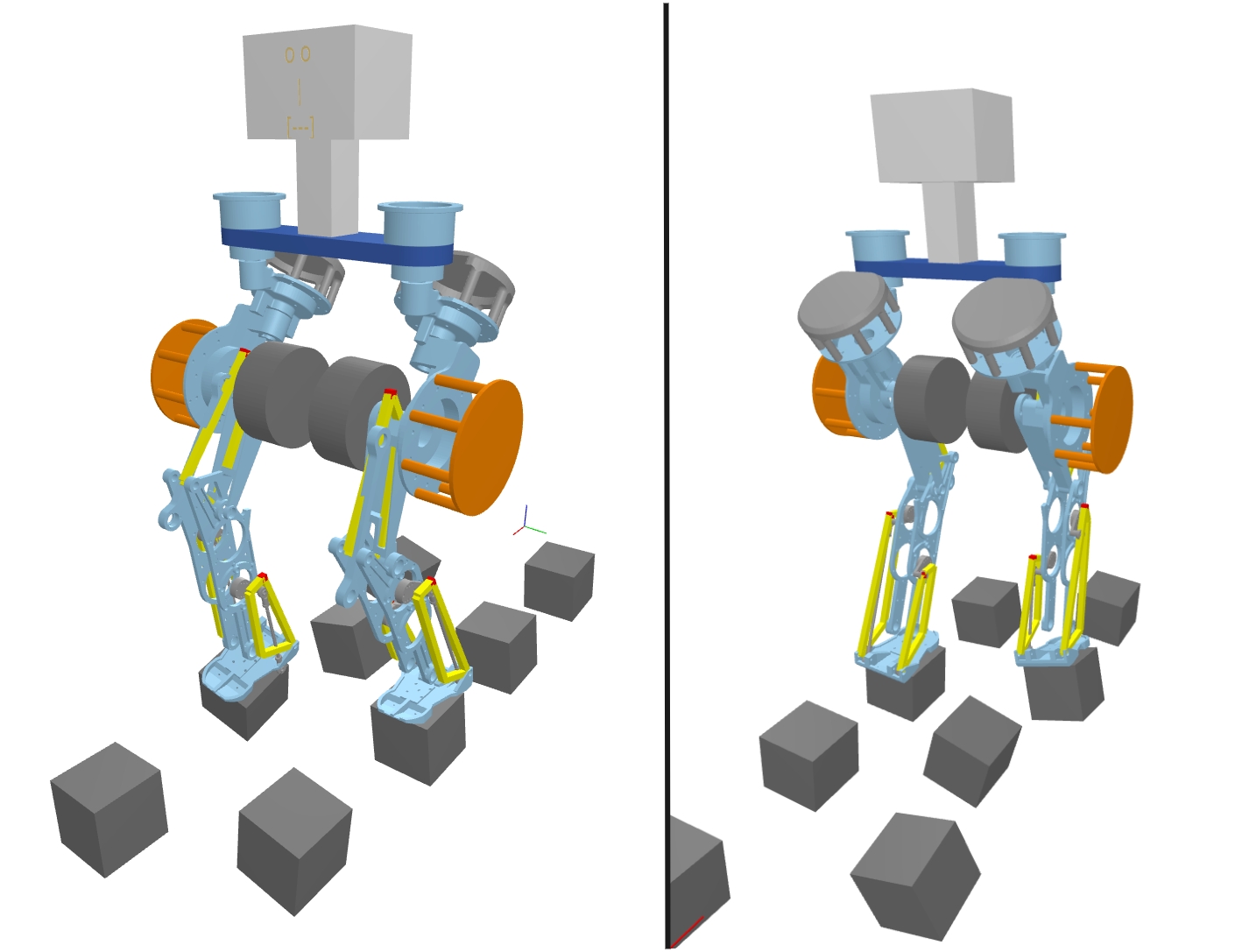}
    \caption{Illustration of the task of walking on rough terrain. Ground orientation is randomly generated for each step.}
    \label{fig:rough_terrain_walking}
\end{figure}

We first perform various tasks to assess the applicability of our \textit{Actuated Serial Model} and demonstrate the limits of the \textit{Minimal Serial Model}.
We perform a walk on a flat terrain at a constant target Center of Mass velocity of 0.7m/s (equivalent to 8km/h walk for a human-sized robot).
On this movement, the \textit{Actuated Serial Model}, that accounts for the closed-loop transmission in the optimization, and the \textit{Minimal Serial} model, that optimizes on serial joints torques then converted into motor controls, yield similar results.\\
However, the two models start to diverge when advancing toward more demanding movements such as walking on rough terrain (see Fig.~\ref{fig:rough_terrain_walking}) and climbing stairs.
\begin{figure}
    \centering
    \includegraphics[width=0.8\linewidth]{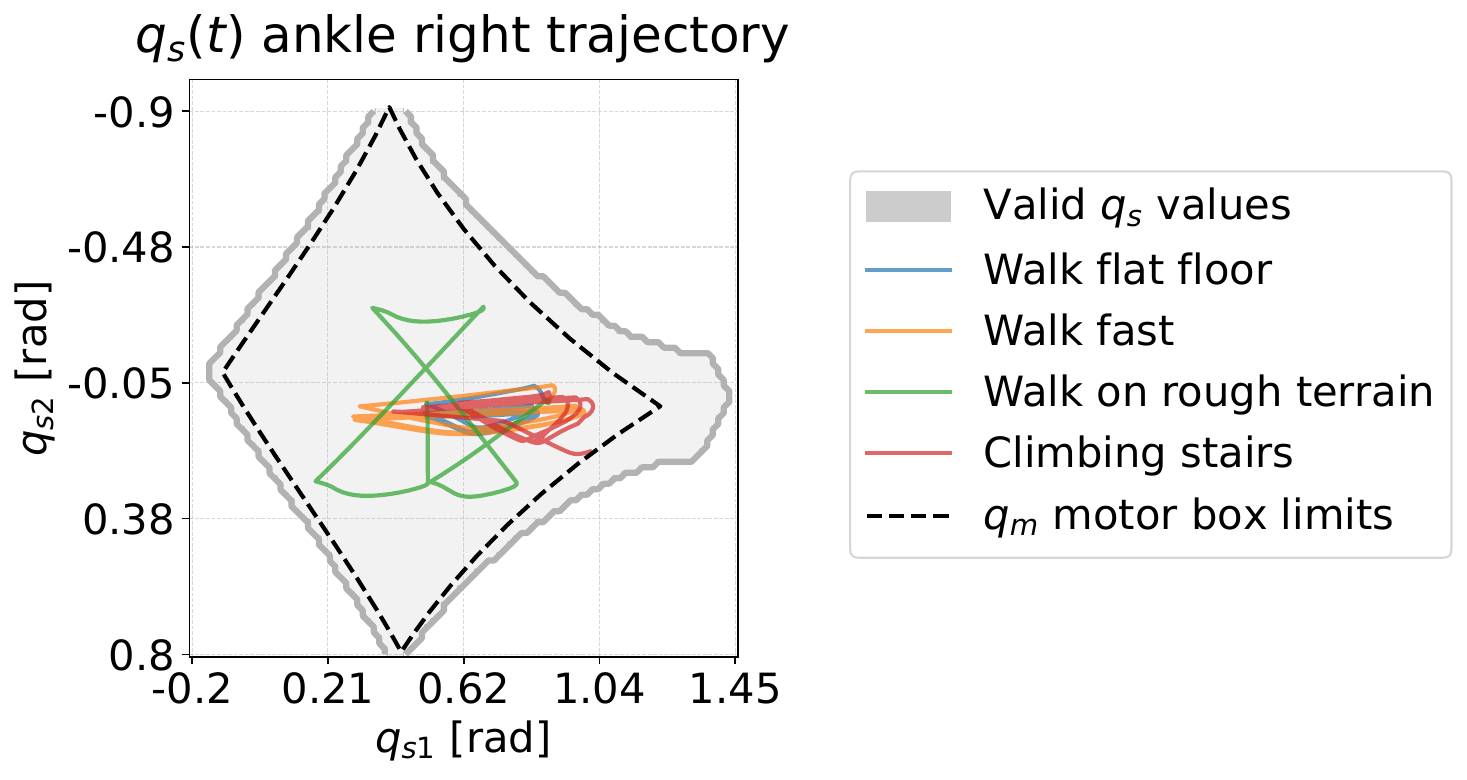}
    \caption{Ankle $q_s(t)$ for walking on flat floor (slowly and fast), walking on rough terrain, climbing stairs. The dotted black lines limit the feasible configurations on the real robot. No serial box constraint can cover all motions while staying in the motor constraints.}
    \label{q_s trajectories}
\end{figure}
Fig.~\ref{q_s trajectories} presents the trajectories of the left and right ankles serial joints, with 2 DoF per ankle and opposes it to the limits of the actuation model, discussed in Sec.~\ref{subsec:OCP_description}.
We observe that the \textit{Actuated Serial Model} reaches ankles configurations that could not be attained with box constraints in the joint-space (that would appear as a straight square limit in Fig.~\ref{q_s trajectories}).
Indeed, clamping the serial joints range to allow walking on flat floor would prevent configurations yet necessary to walk on rough terrains or to climb high stairs.
On the opposite, setting bounds that allow all the motions demonstrated here, would also allow unfeasible configurations.
Our approach can successfully exploit this range of motion by stating the limits in the actuator space, resulting in more permissive, yet feasible, serial limits. 
The different movements can be seen in the companion video.

\subsection{Reinforcement Learning}

% \begin{figure}
% \label{fig:Stiffness motor gain}
%     \centering
%     \includegraphics[width=0.7\linewidth]{img/kp_knee_right.png}
%     \caption{Stiffness gain $K_{Pm}$ of the right knee motor during walking policy deployment. Stiffness becomes variable to match the constant stiffness in serial space.}
%     \label{Stiffness}
% \end{figure}

\begin{figure}
    \centering
    \includegraphics[width=0.7\linewidth]{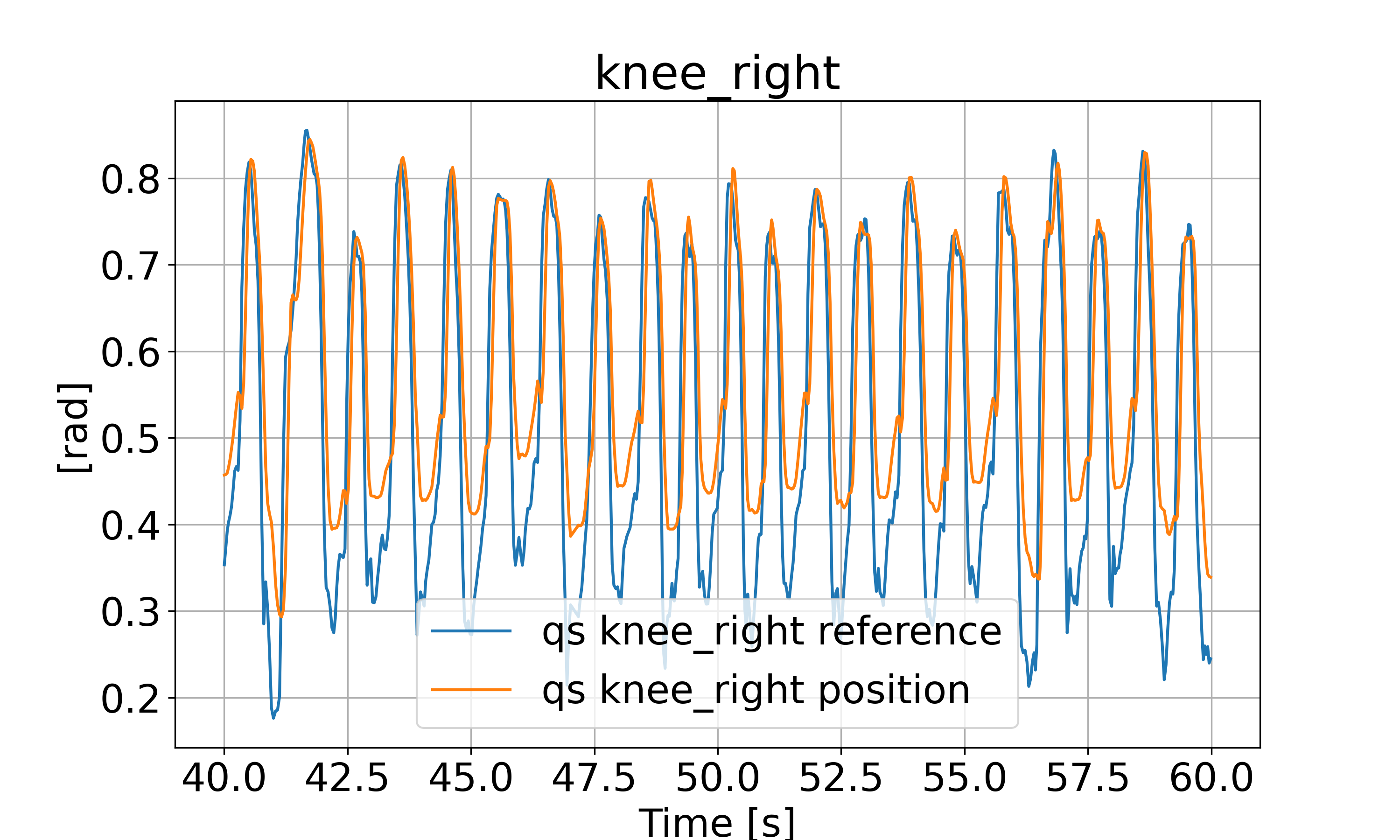}
    \caption{$q_{s}^{*}$ and $q_{s}$ of right knee mechanism during walking policy deployment. Even with $q_{s}^{*}$ outside of the limits, which is frequent in RL, the computed $q_{m}^{*}$, $K_{Pm}$ and $K_{Dm}$ allow to replicate the behavior of the serial joints expected by the policy. }
    \label{policyreference}
\end{figure}

\begin{figure}[h!]
\label{fig:RL_walking}
    \centering
    \includegraphics[width=\linewidth]{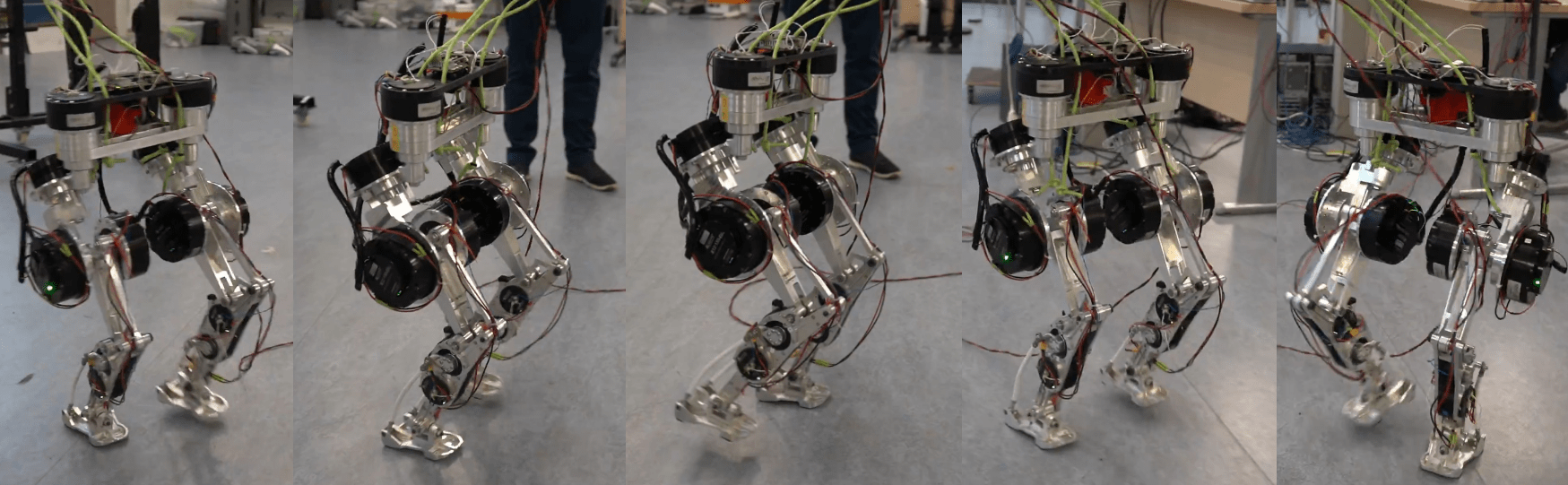}
    \caption{RL with series-parallel impedance transfer allows successful omnidirectional walking deployment on the real robot.}
    \label{snapshot}
\end{figure}

Using the RL implementation presented in Sec.~\ref{sec:implementation}, we were able to deploy a walking policy that was trained on a serial model of the robot.
% respects feasible domains, ensuring not only that the robot configuration remains feasible but also that the motor torques required to generate the movements remain in the limits.
We validated the learned policy in MuJoCo \cite{todorov_mujoco_2012}, to ensure that the policy has not been overfit to Isaac Sim. Our simulation uses a model with the closed-loop mechanisms to validate the complete control strategy.
Our impedance transfer shown in \ref{subsec:gains} and the state estimator shown in \ref{stateestimation} allow successful deployment on the system, with a computation under 10 ms for both, with pure Python code, allowing updates up to 100Hz on a Raspberry PI 5 onboard computer. We leverage the use of the low level control loop of the actuators to stabilize the system, that runs at a way higher rate (typically thousands of Hz). This allows efficient tracking of the serial space reference as showcased in Fig. \ref{policyreference}. To validate our method, we also tested to send the reference torques $\tau_{m}^{*}$ directly to the actuators, which led to dramatic failure, causing the robot to fall.

The walking movements are reported in Fig.~\ref{snapshot} and the companion video. 
This result highlight the potential of the method in handling a wide range of operational scenarios, even beyond restrictive limits on serial joints. Our method allows seamless transfer using the impedance gain conversion, avoiding the drawbacks and difficulties of training with simulated chain closures.

We demonstrated that ankle motor limits play a significant role in more complex motions, such as locomotion on uneven terrain. Nevertheless, these limits were not incorporated into our RL experiments, as our evaluation focuses solely on flat-ground walking. However, our code (which will be released as open source) provides GPU implementations of the models $f$ and $J_A$ using CusAdi \cite{jeon_cusadi_2024}, thereby facilitating the integration of such limits into RL frameworks.

\section{Conclusion}
% In this work, we extended both MPC and RL locomotion frameworks to robots equipped with closed-kinematic actuators, while keeping the computational overhead minimal.
% Our approach builds on a differential geometric description of 2 widespread transmission mechanisms, enabling control of a reduced robot model that still captures transmission effects. We derived the analytical expressions of this actuation model, and its derivatives, making it compatible with derivative-based optimal control solvers. The same derivations allow transferring impedance gains from the serial joint space to the actuator space, thereby enabling the direct deployment of serially trained RL policies on real robots. We validated our method within an optimal control framework through challenging locomotion tasks, demonstrating its ability to handle costs on the full robot state. Moreover, our proposed Actuated Serial Model naturally incorporates richer actuation limits than simplified serial models, thereby broadening the achievable motion repertoire. Finally, we showed the applicability to reinforcement learning by training locomotion policies in serial space and successfully transferring them to actuator space using variable impedance gains that preserve the intended torques.
We presented a framework that extends both MPC and RL locomotion methods to robots equipped with closed-kinematic actuators, while introducing minimal computational overhead. Our approach builds on a differential geometric description of two common transmission mechanisms (knee and ankle), providing analytical expressions of the actuation model and its derivatives. This enables efficient derivative-based trajectory optimization, as well as the transfer of impedance gains from joint space to actuator space, allowing RL policies trained in serial space to be deployed directly on hardware. We validated the method in trajectory optimization on challenging locomotion tasks and demonstrated successful policy transfer to a real robot using the variable impedance gains, which has not been possible without.
We have focused our empirical analysis on explaining and showcasing the importance of modeling the transmission, compared to only modeling the serial kinematics.
Yet the model also offers evident advantages in computational complexity, involving negligible additional computation in addition to the serial model, compared to the extra complexity of modeling the entire constrained kinematics in exhaustive approaches. By delivering practical implementations in both TO and RL, our work highlights how compact transmission models can enhance control accuracy and robustness in modern legged robots, and opens opportunities for future designs with richer non-linear actuation mechanisms.

\bibliographystyle{ieeetr}
\bibliography{main}

% \appendices
% \section{Transfer on the real system} \label{appendix_inverse_mapping}
% In a MPC or RL setting, the actual $q_s$ angles from the robot are needed.
% On the real system, motor encoders provide only $q_m$, requiring the inverse mapping $f^{-1}$ to recover $q_s$.
% For the four-bar mechanism, $f^{-1}$ has an analytical form, but for the more complex ankle mechanism, it does not.
% In both cases, we estimate $q_s$ via numerical optimization:
% \begin{equation} q_s = f^{-1}(q_m) = \min_{q_s} | f(q_s) - q_m |^2 \end{equation}
% State continuity over time provides accurate warm starts, enabling rapid convergence in just a few steps with minimal computation.
\end{document}